\documentclass[default]{sn-jnl}% Default
\usepackage{graphicx}%
\usepackage{multirow}%
\usepackage{amsmath,amssymb,amsfonts}%
\usepackage{amsthm}%
\usepackage{mathrsfs}%
\usepackage[title]{appendix}%
\usepackage{xcolor}%
\usepackage{textcomp}%
\usepackage{manyfoot}%
\usepackage{booktabs}%
\usepackage{algorithm}%
\usepackage{algorithmicx}%
\usepackage{algpseudocode}%
\usepackage{listings}%

\usepackage{booktabs} % For prettier tables
\usepackage{siunitx} % Required for alignment
\theoremstyle{thmstyleone}%
\theoremstyle{thmstyletwo}%

\theoremstyle{thmstylethree}%

\begin{document}

\title[Article Title]{Two Stream Scene Understanding on Graph Embedding}


\author *[1]{\fnm{Wenkai} \sur{Yang}}\email{e0983030@u.nus.edu}

% \author[1]{\fnm{Cheng} \sur{Xiang}}\email{elexc@nus.edu.sg}

\author [2]{\fnm{Wenyuan} \sur{Sun}}\email{e0983371@u.nus.edu}
\equalcont{These authors contributed equally to this work.}

\author [2]{\fnm{Runxaing} \sur{Huang}}\email{e0983021@u.nus.edu}
\equalcont{These authors contributed equally to this work.}

% \author[1,2]{\fnm{Kanglong} \sur{Fan}}\email{iiiauthor@gmail.com}

\affil[1]{ \orgdiv{ECE Department College of Design and Engineer}, \orgname{National University of Singapore}, \orgaddress{\street{E4}, \city{Singapore}, \postcode{100190}, \state{Singapore}, \country{Singapore}}}

\affil[2]{ \orgdiv{Institute of Systems Science}, \orgname{National University of Singapore}, \city{Singapore}, \postcode{119615}, \state{Singapore}, \country{Singapore}}
% \affil[3]{\orgdiv{Department}, \orgname{Organization}, \orgaddress{\street{Street}, \city{City}, \postcode{610101}, \state{State}, \country{Country}}}


\abstract{The paper presents a novel two-stream network architecture for enhancing scene understanding in computer vision. This architecture utilizes a graph feature stream and an image feature stream, aiming to merge the strengths of both modalities for improved performance in image classification and scene graph generation tasks. The graph feature stream network comprises a segmentation structure, scene graph generation, and a graph representation module. The segmentation structure employs the UPSNet architecture with a backbone that can be a residual network, Vit, or Swin Transformer. The scene graph generation component focuses on extracting object labels and neighborhood relationships from the semantic map to create a scene graph. Graph Convolutional Networks (GCN), GraphSAGE, and Graph Attention Networks (GAT) are employed for graph representation, with an emphasis on capturing node features and their interconnections. The image feature stream network, on the other hand, focuses on image classification through the use of Vision Transformer and Swin Transformer models. The two streams are fused using various data fusion methods. This fusion is designed to leverage the complementary strengths of graph-based and image-based features.Experiments conducted on the ADE20K dataset demonstrate the effectiveness of the proposed two-stream network in improving image classification accuracy compared to conventional methods. This research provides a significant contribution to the field of computer vision, particularly in the areas of scene understanding and image classification, by effectively combining graph-based and image-based approaches.}




\maketitle

\section{Introduction}\label{sec1}

Scene understanding is a complex field within computer vision that aims to enable machines to interpret and understand visual data in a manner similar to human perception. The objective is to create systems that can identify and localize individual objects within an image while also comprehending the relationships and functionalities between these objects. Scene graph generation is instrumental for tasks related to scene understanding. 

The generation of scene graphs often relies on scene segmentation, a long-standing and challenging problem in computer vision with numerous downstream applications such as augmented reality, autonomous driving, human-machine interaction, and video content analysis.

The primary motivation for our work in extracting scene graphs from images or videos lies in the graph structure's significant utility for scene understanding. Regardless of image resizing or cropping, the fundamental objects and their relationships within the scene are preserved, making the graph a robust form of representation. To combine features from the graph representation and common image features obtainable through Convolutional Neural Networks (CNN) or Vision Transformer frameworks, we will employ multi-modal methods for data fusion.

Our architecture utilizes a two-stream model for this multi-modal computer vision task. One stream leverages either CNN or Transformer frameworks as the backbone for segmentation, and uses scene graph generation techniques to convert the semantic map into a simplified graph, which is then represented through graph embeddings. The other stream employs conventional strategies, using either the Swin Transformer or ViT (Vision Transformer) for image classification. The fusion methods for combining features from these two streams can be either concatenation or cross-attention mechanisms.

The relationship between objects is crucial for scene understanding, as it is not explicitly captured by existing frameworks. While humans can readily identify the main objects and their relationships, image classification models like CNN or ViT primarily focus on recognizing patterns and pixel-level features in images, rather than understanding the scene as a whole. The two-stream network offers a solution by helping the model comprehend different objects, their positions, and spatial relationships within a scene.

In this paper, we introduce a novel two-stream network architecture leveraging graph representation and Swin Transformer to enhance both image classification and scene graph generation tasks. This advanced scheme transcends conventional image classification approaches that predominantly rely on a single modality, such as CNN or attention mechanisms. The primary contributions of our work are outlined below:
\begin{itemize}
    \item We employ a two-stream network, training each stream independently using the same loss function to predict either the graph or image label. This design facilitates a seamless fusion of different modalities.
    \item In transitioning from segmentation to graph, we have incorporated scene graph generation techniques to elucidate the relationships between neighboring objects effectively.
    \item We have implemented diverse data fusion strategies to ensure coherent alignment of graph and image features.
\end{itemize}

\section{Related Work}\label{sec2}

The scene graph generation has been emerged as a linchpin in enhancing machine understanding of visual scenes by images into structured graphical representations. Xu.\cite{Xu_2017_CVPR} was primarily tried on object detection and relationship between every two objects prediction using CNN\cite{Girshick_2015_ICCV}\cite{krizhevsky2012imagenet} and Region Proposal Network(RPN). This work directly utilize the set of object proposals to predict the object categories, bounding box offsets, and relationship types by Gated Recurrent Unit (GRU)\cite{cho2014learning} and graph pooling layer. Alejandro Newell et al.\cite{newell2017pixels} proposed a single-stage CNN for object and relationship detection, transitioning from pixel representations to graph structures through two heatmaps. These heatmaps activate at the predicted locations of objects and relationships, providing a useful representation for various computer vision tasks.

In 2021, attention mechanisms gained traction in computer vision tasks, exemplified by the advent of the Image Transformer\cite{dosovitskiy2010image}. Empirical evidence suggests that, compared to CNNs, Transformers could exhibit superior performance on large datasets, especially when pre-trained. This theoretical advancement proved beneficial for later multimodal works, such as CLIP (Contrastive Language-Image Pre-training)\cite{radford2021learning}, as both computer vision and text processing tasks could leverage the same architectural underpinning to extract corresponding features. Following this trend, the Swin Transformer was introduced\cite{liu2021swin}, which incorporates shifted window attention and local window attention mechanisms, similiar to the operation of CNNs. This innovation broadened the scope of Transformers, making them applicable to other computer vision tasks like segmentation.

The first two stream network has been used in the action recognition in video\cite{simonyan2014two}. Simonyan first propose the two-stream ConvNet architecture which incorporates spatial and temporal networks. The spatial stream network input is the common singe RGB image, while the temporal stream network input is the multi-frame optical flow. To the action video classification, the temporal stream network could highly improve the accuracy and complement the lacking modality information. Actually after the work, there is work\cite{gammulle2017two} combine the LSTM\cite{sak2014long} and CNN to sovle the video classification. But there is some difference between theses two works, the Simonyan's work use two parallel two stream networks, however the the gammulle directly use CNN extract every frame feature and feed them as a sequence into the LSTM network.

The Kipf\cite{kipf2016semi} introduce the graph convolutional operator to do Semi-supervised graph classification, and the hamilton\cite{hamilton2017inductive} utilize inductive framework that leverages node feature information (eg, text attributes) to efficiently generate node embeddings. After the transformer\cite{vaswani2017attention} appeared, the velivckovic \cite{velivckovic2017graph} applied the attention on the graph-structured data. Generally we have tried some different data fusion like late fusion\cite{snoek2005early}, and the different fusion methods have been adapted such as concatenation, cross attention, voting, averaging and Canonical Correlation Analysis (CCA)\cite{hardoon2004canonical}.

The contrastive learning\cite{chen2020simclr} is a powerful self-supervised learning technique. The primary goal is to learn a common representation space where similar items from different modalities are close to each other, while dissimilar items are far apart. In CLIP, this is achieved by training a neural network to map images and text descriptions into a shared embedding space. Utilize a contrastive loss function, such as the Noise Contrastive Estimation (NCE)\cite{gutmann2010noise} loss or InfoNCE\cite{oord2018representation} loss, to train the model. The loss function encourages the model to minimize the distance between positive pairs and maximize the distance between negative pairs in the shared embedding space. In our case, the positive pair is the same image of graph embedding and pixel embedding, and the negative pair is the different image of graph embedding and pixel embedding.

\section{Methods}\label{sec3}

\subsection{Graph Feature Stream Network}

\subsubsection{The Segmentation Structure}
\begin{figure}[h]
    \begin{center}
    \includegraphics[width=\linewidth]{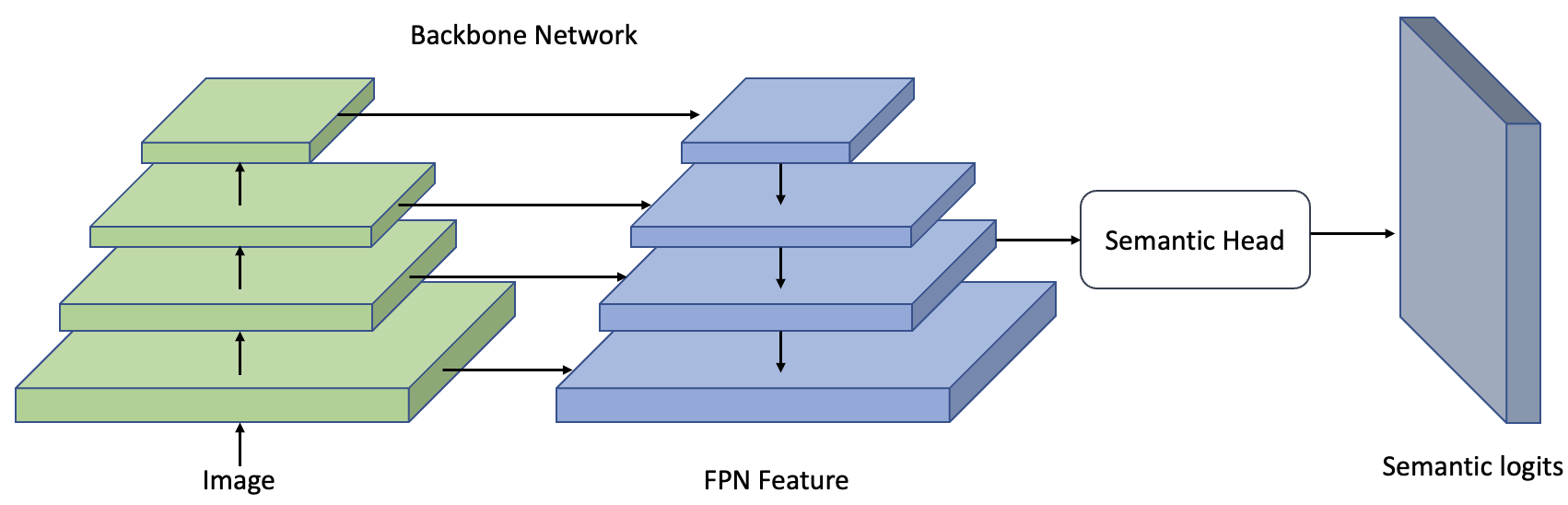}
    \caption{The architecture of Segmentation}
    \label{fig:segmentic}
    \end{center}
\end{figure}

In segmentation part, we use the Upsnet\cite{xiong2019upsnet} as whole architecture, which is a unified panoptic segmentation network for tackling the newly proposed segmentation task. The backbone could be residual network, Vision Transformer(Vit)\cite{dosovitskiy2020image} and Swin Transformer. The deformable convolution based semantic segmentation head and a Mask R-CNN style instance segmentation head have been used to solve these two subtasks(Semantic Segmentation and Instance Segmentation) simultaneously. We adopt the original Mask R-CNN\cite{he2017mask} backbone and Swin Transformer as our feature extraction network. The Mask R-CNN backbone exploits a deep residual network\cite{he2016deep} (ResNet) with a feature pyramid network\cite{lin2017feature} (FPN). The architecture is in Fig.\ref{fig:segmentic}

\begin{figure}[h]
    \begin{center}
    \includegraphics[width=\linewidth]{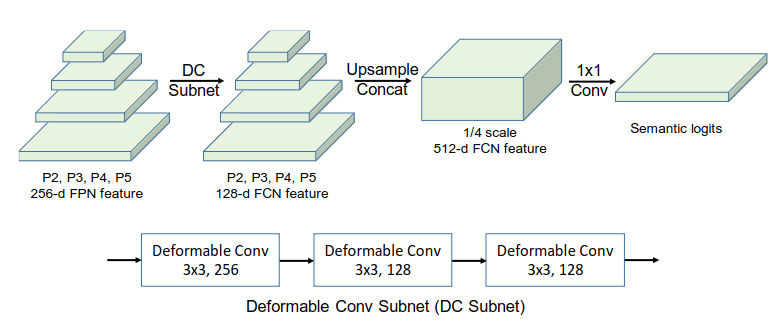}
    \caption{The architecture of semantic segmentation head}
    \label{fig:segmentic head}
    \end{center}
\end{figure}

The goal of the semantic segmentation head is to segment all semantic classes, Our semantic head consists of the multi-scale feature from FPN as input. In particular, we use P2, P3, P4 and P5 feature maps of FPN which contain 256 channels and are 1/4, 1/8, 1/16 and 1/32 of the original scale respectively. These feature maps first go through a deformable convolution network\cite{dai2017deformable} independently and are subsequently upsampled to the 1/4 scale. We then concatenate them and apply 1 × 1 convolutions with softmax to predict the semantic class. The architecture is shown in Fig. \ref{fig:segmentic head}. Semantic segmentation head is associated with the regular pixel-wise cross entropy loss. To put more emphasis on the foreground objects such as pedestrians, the RoI loss is also incorporated. 

Pixel-wise cross entropy is often used for semantic segmentation tasks where each pixel of an image is classified into one of $C$ classes. The loss for each pixel is computed as the cross entropy between the predicted probabilities and the true class. Given that $p_c(i,j)$ is the predicted probability of pixel at position $(i,j)$ belonging to class $c$ and $y_c(i,j)$ is the ground truth label (1 if the pixel belongs to class $c$, 0 otherwise), the pixel-wise cross entropy loss $L$ can be given as:

\begin{equation}
    L = -\sum_{i,j}\sum_{c=1}^{C} y_{c}(i,j) \log(p_{c}(i,j))
\end{equation}

\begin{figure}[h]
    \begin{center}
    \includegraphics[width=\linewidth]{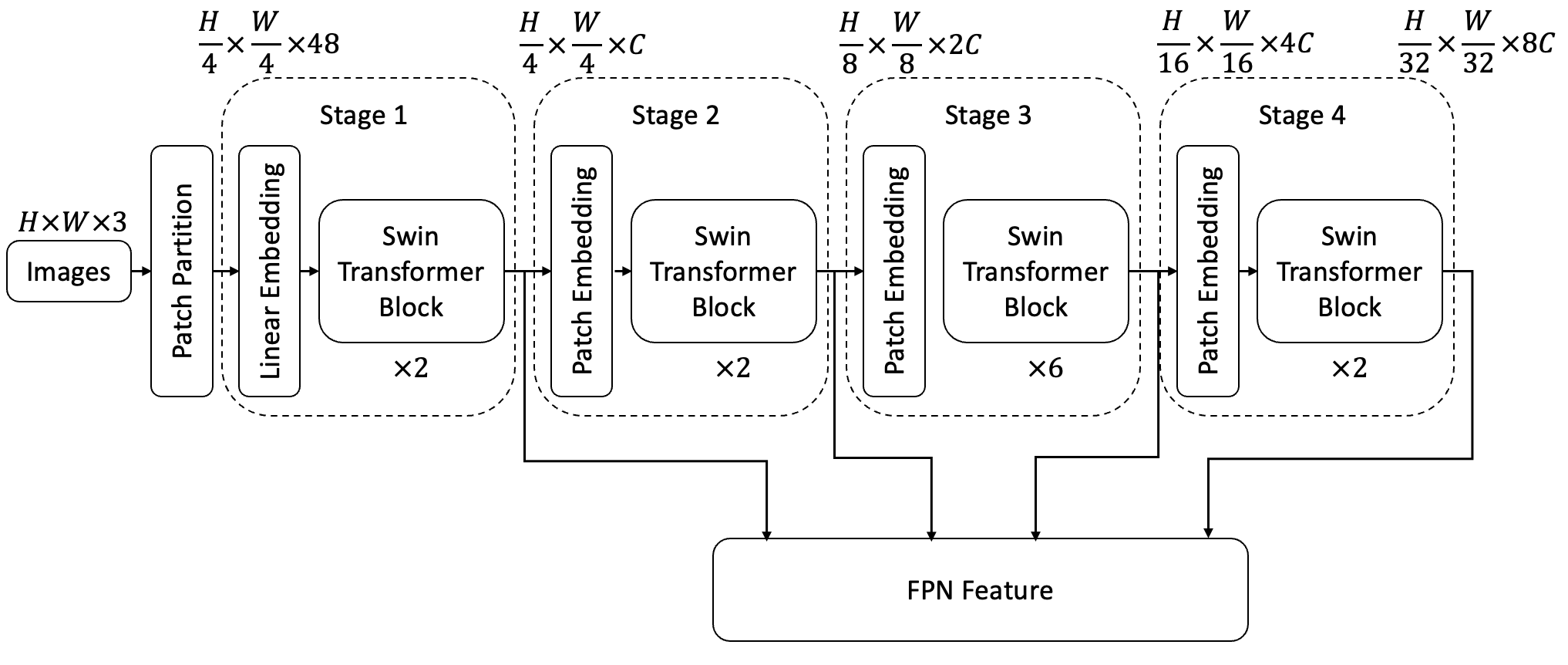}
    \caption{The architecture of Swin Transformer}
    \label{fig:Swin}
    \end{center}
\end{figure}

We use the Swin Transformer as the backbone in UPSNet. Swin Transformer divides an input image into non-overlapping patches, linearly embeds them, and then feeds them into a series of Transformer layers. Unlike conventional Transformers, the Swin Transformer uses shifted windows to allow local features to be aggregated over layers. This results in multiple levels of feature resolutions throughout its layers, similar to the hierarchical features in a CNN. The swin architecture with FPN is shown in Fig. \ref{fig:Swin}.

\subsubsection{The Scene Graph Generation}
\begin{figure}[h]
    \begin{center}
    \includegraphics[width=\linewidth]{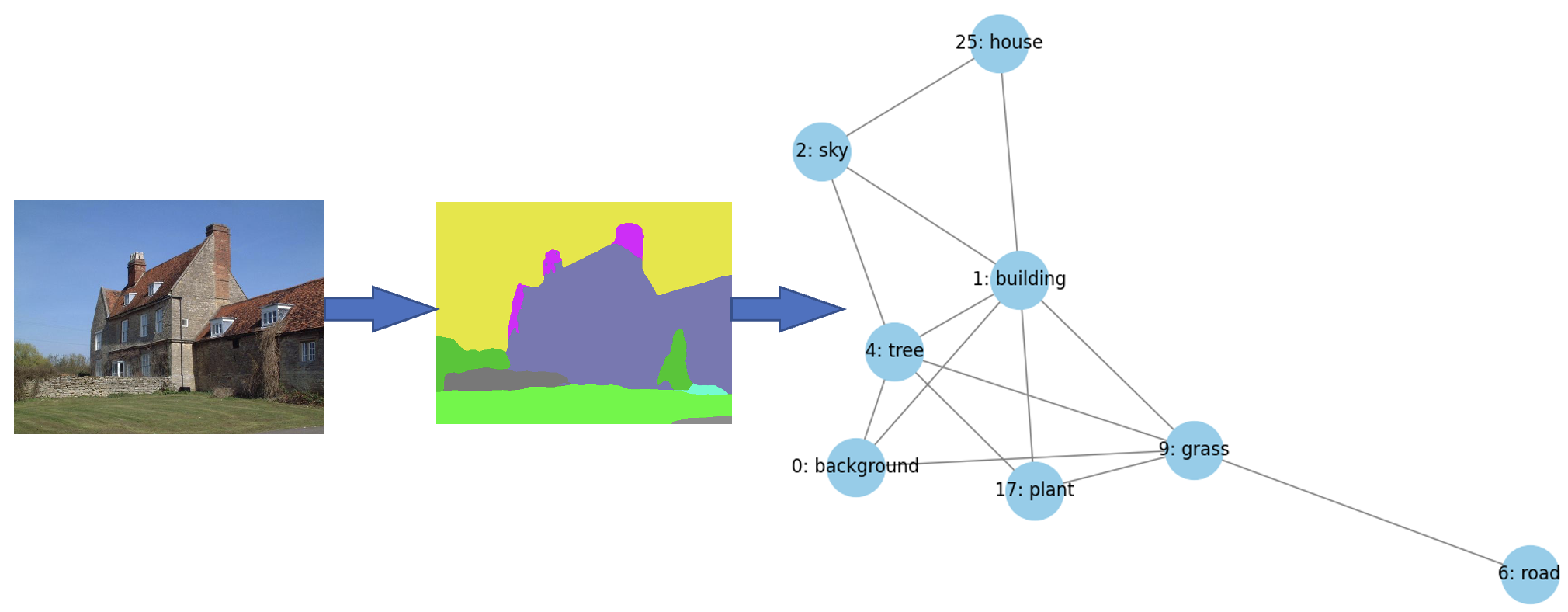}
    \caption{The example of Scene Graph Generation}
    \label{fig:Scene}
    \end{center}
\end{figure}

The success of segmentation or detection has surged interest in examining the detailed structures of a visual scene, especially in the form of object relationships. Scene graph offers a platform to explicitly model objects and their relationships. In short, a scene graph is a visually-grounded graph over the object instances in an image, where the edges depict their pairwise relationships, see example in Fig.\ref{fig:Scene}. In this work, we address the problem of scene graph generation, where the goal is to generate a visually-grounded scene graph from an image segmentation. 

\begin{figure}[h]
    \begin{center}
    \includegraphics[width=\linewidth]{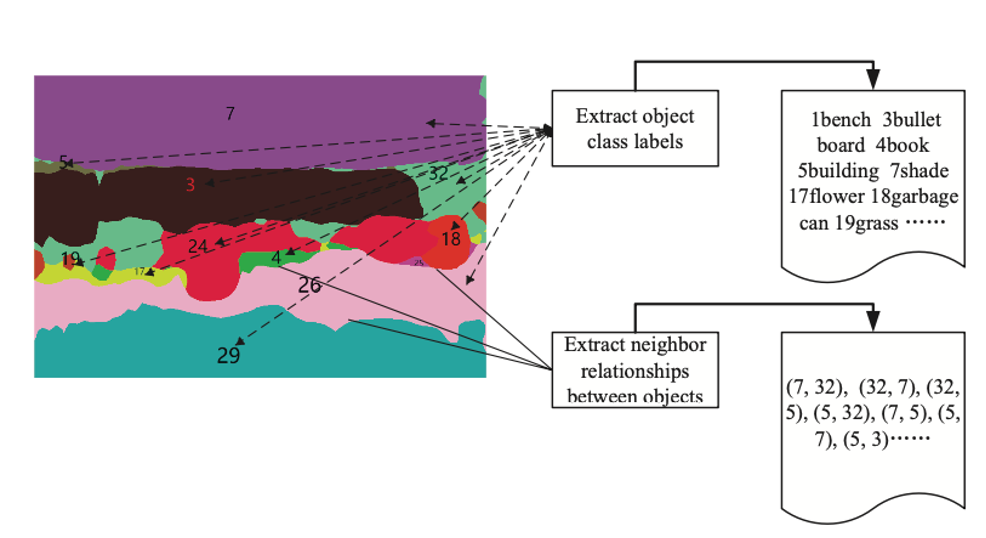}
    \caption{The illustration of generating scene graph}
    \label{fig:seg}
    \end{center}
\end{figure}

From the semantic map, object labels can be extracted and according to the boundary between different labels, a neighborhood relationship between objects can be established. Therefore, we need to detect the boundary of pixels in the semantic map row by row and column by column, so as to get the possible neighborhood relationship between different objects of the original image, as shown in Fig. \ref{fig:seg}. Depending on the object category labels and the neighborhood relationship between different objects in original image, we can obtain the scene graph.
First, we convert the labels of objects, which is provided by the semantic map consisting of $C$ classes into the nodes in scene graph. The node feature could be represented by the vector($C$ length) via one hot encoding. The neighborhood relationship is determined simply by the presence or absence of pixel intersection and could be regraded as the edge(the pair of two nodes) in scene graph. 

\subsubsection{The Graph Representation}

Graph Neural Network(GNN) recently has received a lot of attention due to its ability to analyze graph structural data. A set of objects, and the connections between them, are naturally expressed as a graph. There are three general types of prediction tasks on graphs: graph-level, node-level, and edge-level. In a graph-level task, we predict a single property for a whole graph. For a node-level task, we predict some property for each node in a graph. For an edge-level task, we want to predict the property or presence of edges in a graph. In this work, we will predict the label of graph means the corresponding image, because we intend to get the feature of graph.

Graph Convolutional Network(GCN)\cite{kipf2016semi} layer is one of the pioneering methods in GNNs. The key idea of a GCN layer is to update the representation of a node by aggregating information from its neighbors.

Given:
\begin{itemize}
    \item $A$: adjacency matrix of the graph
    \item $X$: node features matrix
    \item $H^{(l)}$ : node representations at layer $l$
    \item $W^{(l)}$: weight matrix at layer $l$
\end{itemize}
The basic update rule for a GCN layer is:
\begin{equation}
H^{(l+1)} = \sigma\left( \tilde{D}^{-\frac{1}{2}} \tilde{A} \tilde{D}^{-\frac{1}{2}} H^{(l)} W^{(l)} \right)
\end{equation}
Where:
\begin{itemize}
    \item $\tilde{A} = A + I_N$ is the adjacency matrix of the undirected graph $G$ with added self-connections, $I_N$ is the identity matrix.
    \item $\tilde{D}_{ii} = \sum_j \tilde{A}_{ij}$ and $W^{(l)}$ is a layer-specific trainable weight matrix.
    \item $\sigma$ is a non-linear activation function, e.g. ReLU.
\end{itemize}
GCN uses a fixed-weight averaging scheme based on the adjacency matrix of the graph.

GraphSAGE (Graph Sample and Aggregation)\cite{hamilton2017inductive} is another popular GNN method, which is designed to generate embeddings by sampling and aggregating features from a node's neighbors.

The key steps in a GraphSAGE layer are:
\begin{itemize}
    \item Neighbor Sampling: Sample a fixed-size set of neighbors for each node.
    \item Feature Aggregation: Aggregate the features of the sampled neighbors.
    \item Combination: Combine a node's current features with the aggregated neighborhood features.
\end{itemize}

The aggregation function can be mean, LSTM, pooling, etc. For the mean aggregator, the update rule can be given as:

\begin{equation}
h_{\text{neigh}(v)}^{(l)} = \text{MEAN}\left( \{ h_u^{(l-1)} : u \in N(v) \} \right)
\end{equation}
\begin{equation}
h_v^{(l)} = \sigma\left( W \cdot \text{CONCAT}\left( h_v^{(l-1)}, h_{\text{neigh}(v)}^{(l)} \right) \right)
\end{equation}
Where:
\begin{itemize}
    \item $h_v^{(l-1)}$ is the representation of node $v$ at layer $l$.
    \item $N(v)$ is the set of neighbors of node $v$.
\end{itemize}
GraphSAGE uses neighborhood sampling and various aggregation functions to update node embeddings.

GAT (Graph Attention Network)\cite{velivckovic2017graph} introduces the concept of attention mechanisms to the graph domain. The main idea is to assign different attention weights to different neighbors of a node, indicating the importance of each neighbor's information.

Given:
\begin{itemize}
    \item $W$: weight matrix
    \item $a$: attention mechanism's weight
    \item $h_i$ and $h_j$ : node feature representations
\end{itemize}
The attention coefficient between two nodes ii and jj can be computed as:
\begin{equation}
e_{ij} = \text{LeakyReLU}\left( a^T [W h_i || W h_j] \right)
\end{equation}
Where $\|$ denotes concatenation. These coefficients are then normalized across all choices of $j$ using a softmax:
\begin{equation}
\alpha_{ij} = \frac{\exp(e_{ij})}{\sum_{k \in N(i)} \exp(e_{ik})}
\end{equation}
The new node representation can then be computed as a weighted sum of its neighbors:
\begin{equation}
h_i' = \sigma\left( \sum_{j \in N(i)} \alpha_{ij} W h_j \right)
\end{equation}
GAT employs attention mechanisms to weigh the importance of each neighbor during aggregation.

\begin{figure}[h]
    \begin{center}
    \includegraphics[width=\linewidth]{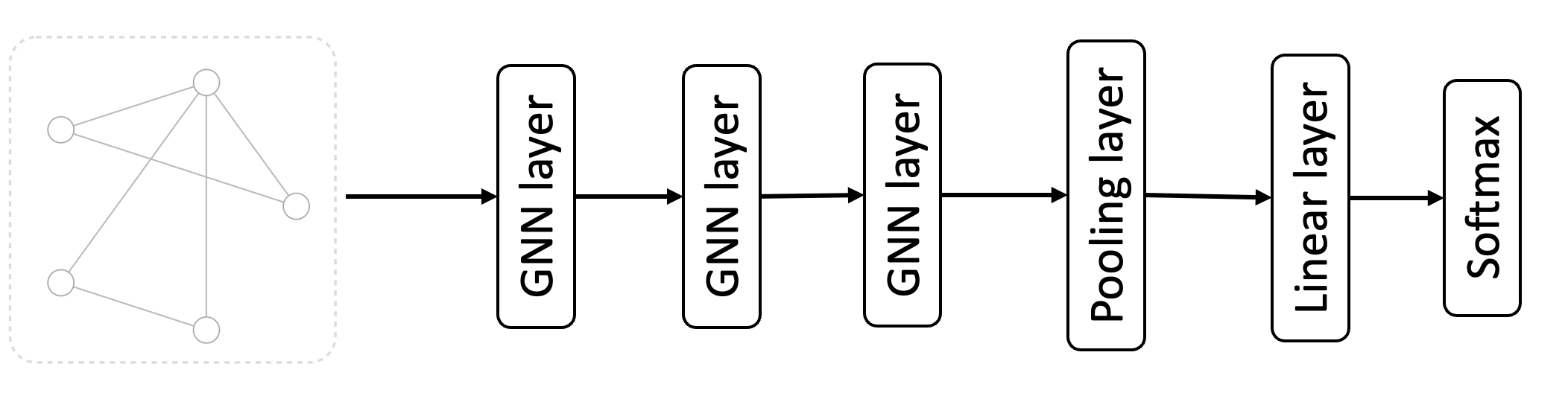}
    \caption{The architecture of generating scene graph}
    \label{fig:gnn}
    \end{center}
\end{figure}

The graph will pass three GNN Layers where each layer performs node feature propagation aggregating information from its neighbor nodes and follow a Readout Layer which could aggregate information from all nodes in a graph to produce a single vector representation of the entire graph, shown in Fig. \ref{fig:gnn}. Common readout functions include Sum, Mean and Max-pooling. 

\subsection{Image Feature Stream Network}
\subsubsection{The Classification Net}
\begin{figure}[h]
    \begin{center}
    \includegraphics[width=\linewidth]{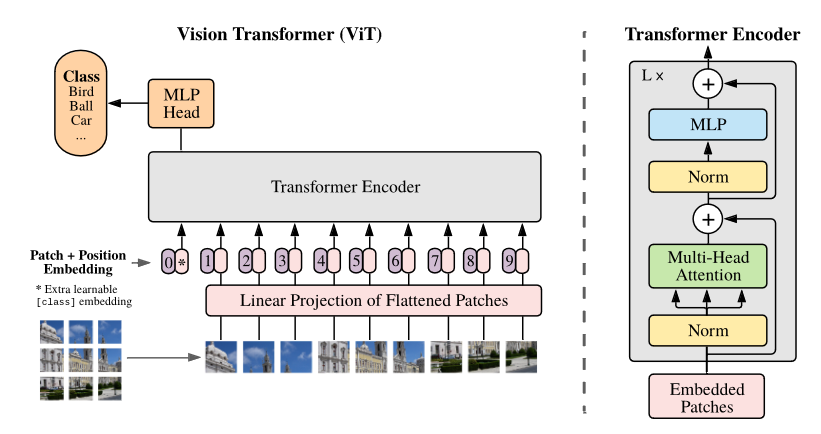}
    \caption{The architecture of Vision Transformer}
    \label{fig:vit}
    \end{center}
\end{figure}

The image classification is a common computer vision task which could be solved by Conv or Transformer. In this section, we will try to use Vision Transformer shown in Fig.\ref{fig:vit} and Swin Transformer shown in Fig.\ref{fig:Swin} to do image classification. 

Vit utilize self-attention mechanisms, originally used in NLP tasks, to capture global dependencies within images. The Vision Transformer treats an image as a sequence of fixed-size patches and processes these patches through a series of transformer blocks. Divide the image into fixed-size patches (e.g., 16x16 pixels). Flatten each patch and linearly embed each of them with a trainable linear projection to create patch embeddings. Optionally add positional embeddings to retain positional information. Pass the sequence of patch embeddings through multiple layers of transformer encoders. Each encoder layer includes multi-head self-attention and MLP blocks. 

Swin introduces shifted windows, which limit self-attention computation to non-overlapping local windows while also allowing for cross-window connections. The preprocessing is same as ViT, but with potentially different patch sizes. The image is divided into non-overlapping windows. Within each window, self-attention is computed (W-MSA - Window based Multi-head Self Attention). Every alternate block shifts the windows to allow for cross-window connection (SW-MSA - Shifted Window based Multi-head Self Attention). Adjacent patches are merged progressively to reduce the number of tokens and increase the receptive field. The final layer's feature map is used for classification, often through a global average pooling layer followed by a fully connected layer (classifier).

In this part we will use the pre-trained model on ImageNet-21k\cite{deng2009imagenet} and full parameter fine-tuning on our ADE20K dataset.

The image feature we need is the cls token after the Transformer Encoder in Vision Transformer or the feature map after stage 4 Swin Transformer Block.

\subsection{The Two Stream Network}

The two-stream network operates on separate training pathways, where the objective of the loss function is to accurately predict the image's label. The graph network is designed to distill object features and their interrelations within the image, effectively mapping out a graph-like representation of the scene. Concurrently, the image classification network applies conventional methods to categorize the image.

Our aim is to extract features that remain consistent regardless of image transformations such as resizing, cropping, or rotation—mirroring the human capability to interpret a scene by discerning the constituent objects and their spatial relationships. We posit that these features closely align with a spatial graph representation.

\begin{figure}[h]
    \begin{center}
    \includegraphics[width=\linewidth]{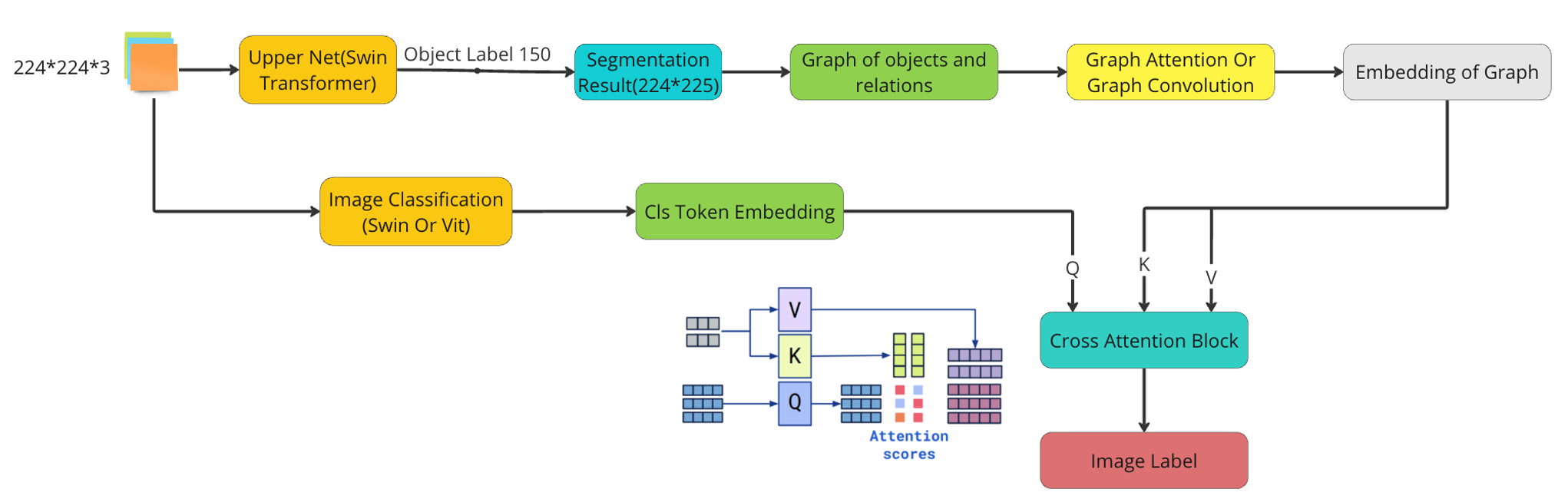}
    \caption{The architecture of Two Stream Network}
    \label{fig:two}
    \end{center}
\end{figure}

By combining graph-based features with those derived from image analysis, we anticipate a bolstered performance in image classification and related tasks. This fusion strategy leverages the complementary strengths of both feature sets to enhance the model's robustness and interpretive accuracy.The whole two stream network shown in Fig.\ref{fig:two}

\subsubsection{The Features Fusion}
The common data fusion methods are concatenation, sum, average, product and voting. In this paper we utilize the cross attention mechanism\cite{Huang_2019_ICCV}\cite{wei2020multi} to do the late fusion. The attention mechanism is often use in the multi-modility work. Cross-attention is a mechanism often found in transformer architectures that allows the model to weigh the importance of different parts of input data from one source when processing another. It's a type of attention that enables the model to attend across different modalities or feature sets, making it particularly useful for tasks that involve multiple types of data (e.g., text and images, audio and video).

Cross-attention can be employed to allow one modality to influence the processing of another. When you have two different modality vectors, such as an image vector and a graph representation vector, cross-attention can be used to fuse these vectors by allowing each modality to attend to and influence the representation of the other.

Implement a cross-attention layer that takes the encoded vectors from both modalities. In this layer, one modality serves as the "query" while the other serves as the "key" and "value" (following the standard transformer nomenclature), shown in Fig. \ref{fig:two}. Queries, keys, and values are projected into a shared space where the compatibility of the query with each key is computed, often using a dot-product attention mechanism. The output is a weighted sum of the value vectors, with weights corresponding to the computed attention scores. 

After cross-attention, the updated vectors that now contain information from both modalities can be used in the corresponding head layer in different tasks. In classification task, we will use a fully connected layer to do projection from the cross-attention output dimension to the classes size.

\section{Experiments}

\subsection{Dataset}
\begin{figure}[h]
    \begin{center}
    \includegraphics[width=\linewidth]{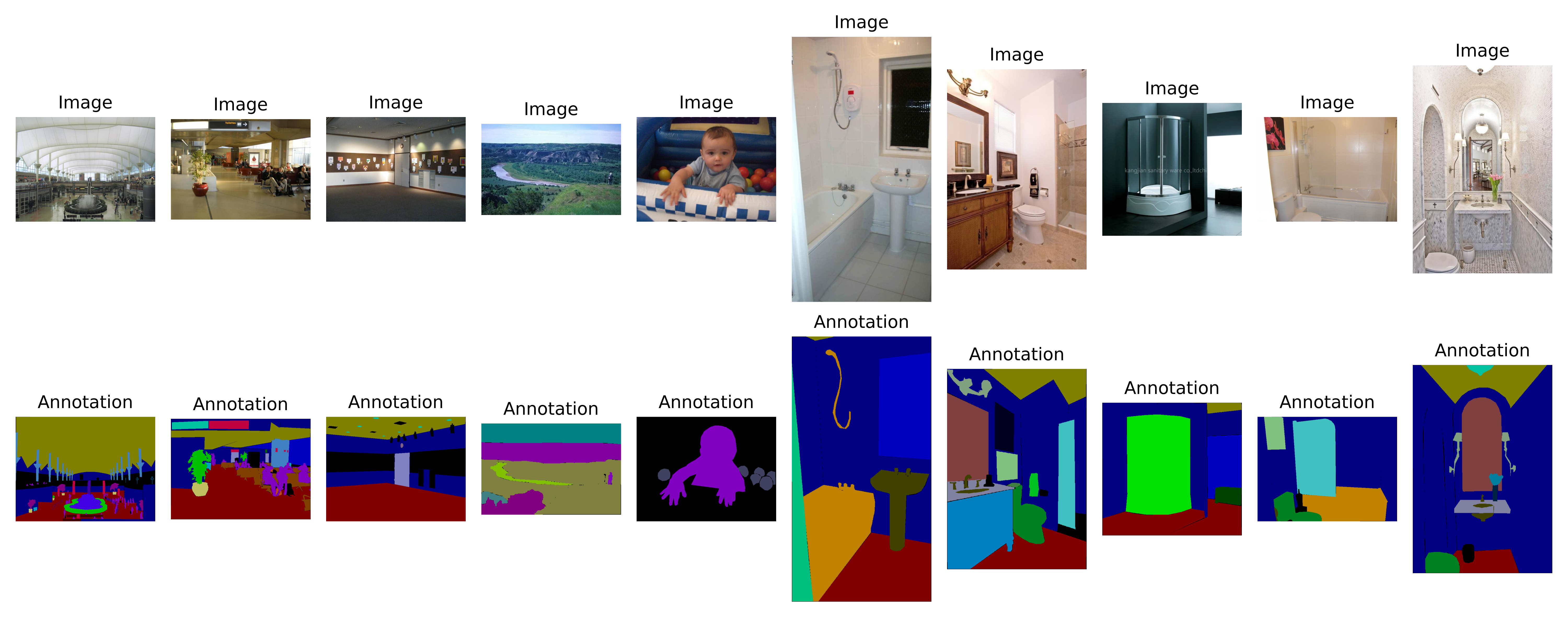}
    \caption{The overview of ADE20k Dataset}
    \label{fig:dataset}
    \end{center}
\end{figure}

The ADE20K\cite{zhou2017scene} dataset is a widely used benchmark in the field of computer vision, particularly in the context of semantic segmentation. In deep learning research, it is widely used in segmentation task. ADE20K is an openly available dataset, making it convenient for comparing and evaluating the performance of various algorithms and models. This dataset comprises a diverse collection of images with pixel-level annotations, scene or image label, covering a broad spectrum of scenes and objects. With over 20,000 images and more than 150 object categories(grass, sky, building etc.), and 1055 scecne categories(airport, basketball court etc.), ADE20K provides a rich and varied source of data for training and testing deep learning algorithms, shown in Fig. \ref{fig:dataset}. The annotated images cover the scene categories from the SUN and Places database. Here there are some examples showing the images, object segmentations, and parts segmentations. It enable you to explore and develop advanced models to achieve state-of-the-art results and potentially contribute to the publication of impressive research papers in this domain.

\subsection{Model Settings}

\textbf{The Segmentation Net}: We conduct experiments on the ADE20K dataset using Swin-L, Swin-B, Convnext-L and Convnext-B which are pre-trained on the ImageNet-21k. The image augmentations have been applied, including normalizing, resize and crop, which could let the image has standard size(512, 512). More details on Table. \ref{tab:training_settings}

\begin{table}[h!]
  \centering
  \begin{tabular}{@{}lllS[table-format=1.2e-1]lS[table-format=3]@{}}
    \toprule
    {Optimizer} & {Learning Rate} & {Weight Decay} & {Batch Size} & {Hardware} & {Epochs} \\
    \midrule
    SGD         & 1e-6            & 5e-7           & 6           & RTX 3090        & 20      \\
    Adam        & 6e-5            & 1e-6           & 6           & RTX 3090       & 20       \\
    RMSprop     & 5e-4            & 0              & 6          & RTX 3090        & 20      \\
    \bottomrule
  \end{tabular}
  \caption{Segmentation model training settings}
  \label{tab:training_settings}
\end{table}

\textbf{Graph Representation}: First we pack the scene graph dataset, and try the different graph neural net to compare the accuracy of graph(image) classification. Training details on 

\begin{table}[h!]
  \centering
  \begin{tabular}{@{}lllS[table-format=1.2e-1]lS[table-format=3]@{}}
    \toprule
    {Optimizer} & {Learning Rate} & {Weight Decay} & {Batch Size} & {Hardware} & {Epochs} \\
    \midrule
    SGD         & 1e-4            & 5e-5           & 256           & RTX 3090        & 150      \\
    Adam        & 1e-4            & 5e-5           & 256           & RTX 3090       & 150       \\
    \bottomrule
  \end{tabular}
  \caption{GNN model training settings}
  \label{tab:training_settings}
\end{table}

\subsection{Result}

\textbf{The Segmentation Net}: Table.\ref{tab:model_results} show the different backbone segmentation model metrics on the ADE20K dataset. 

\begin{table}[h!]
  \centering
  \begin{tabular}{@{}cccccc@{}}
    \toprule
    {Backbone} & {Validation mean iou} & {Overall pixel accuracy} & {Mean pixel accuracy} & {Parameters} \\
    \midrule
    Swin-L         & 48\%            & 89\%            & 74\%            & 197M  \\
    Swin-B        & 45\%            & 82\%            & 69\%            & 88M   \\
    Convnext-L    & 51\%            & 92\%            & 81\%            & 198M   \\
    Convnext-B    & 48\%            & 90\%            & 77\%            & 89M   \\
    \bottomrule
  \end{tabular}
  \caption{Segmentation model results}
  \label{tab:model_results}
\end{table}

\textbf{Graph Representation}: Table.\ref{tab:gnn_results} presents the metrics for various GNN models applied to our graph dataset. The results from the graph representation show a significant improvement in scene understanding. This confirms the essential role of graph modality in enabling the network to recognize scenes. Additionally, the object neighbor feature enhances the performance of two stream networks and facilitates data fusion.

\begin{table}[h!]
  \centering
  \begin{tabular}{@{}cccccc@{}}
    \toprule
    {Network} & {Train accuracy} & {Test accuracy}  \\
    \midrule
    GCNConv                & 63\%            & 48\%         \\
    SAGE conv graph        & 48\%            & 46\%         \\
    Graph Attention        & 43\%            & 43\%         \\
    \bottomrule
  \end{tabular}
  \caption{Graph Representation model results}
  \label{tab:gnn_results}
\end{table}

\textbf{Two stream network}: Our two-stream network task focuses on image classification. As shown in Table.\ref{tab:two_stream}, there is a notable improvement when compared to conventional image classification methods. Our experiments demonstrate that incorporating scene graphs and graph representations can enhance these common methods. From another perspective, the two-stream network architecture effectively fuses two different modalities or features through independent training.

\begin{table}[h!]
  \centering
  \begin{tabular}{@{}cccccc@{}}
    \toprule
    {Network} & {Accuracy} & {Parameters}  \\
    \midrule
    Swin-L   & 69.3\%            & 197M         \\
    Vit-L    & 66.8\%            & 304M         \\
    Our Two Stream Network(cross-attention fusion)   & 71.2\%            & 278M          \\
    Our Two Stream Network(concatenation fusion)        & 70.2\%            & 278M         \\
    \bottomrule
  \end{tabular}
  \caption{Two stream network results}
  \label{tab:two_stream}
\end{table}

\section{Conclusion}

In this paper, we introduce a novel two-stream network capable of generating a scene graph from a segmentation map and classifying images based on a fused graph representation. Our approach involves training a transformer-based scene segmentation model to categorize pixel objects. Concurrently, we develop an automated algorithm to determine neighborhood relationships and create the corresponding scene graph. We employ various Graph Neural Networks (GNN) for graph classification, aimed at deriving graph representation features.

The second stream of our network is a standard image classification model, which incorporates transformer encoding and a classifier head. The key of our approach lies in merging features extracted from both streams: the graph features and the image features. We experimented with several data fusion methods, including cross-attention, concatenation, sum, average, product, and voting.

Our experimental results, particularly on the ADE20K dataset, demonstrate significant improvements with our two-stream network. This indicates that our scene graph representation and integration methods can revolutionize traditional image classification approaches. Furthermore, in considering the graph feature as an intermediary form of information, we open new possibilities for image encoders.

Looking ahead, we plan to extend our two-stream network to other applications and datasets, such as object detection and classification of object relationships.

\bibliographystyle{unsrt}
\bibliography{references.bib}

\end{document}